%% file: paper_main.tex
\documentclass[11pt,a4paper]{article}
\usepackage{emnlp-ijcnlp-2019}
\usepackage{times}
\usepackage{latexsym}
\usepackage{multirow}
\usepackage{booktabs} 
\usepackage{url}
\usepackage{amsmath,amssymb,amsfonts}
\usepackage{graphicx}
\usepackage{graphics}
\usepackage{enumitem} 

\usepackage{algorithm,algpseudocode}
\aclfinalcopy

\title{Model Compression with Multi-Task Knowledge Distillation for Web-scale Question Answering System}

\author{Ze Yang~$^\S$\thanks{\ \ Contribution during internship at STCA NLP Group, Microsoft, Beijing, China.} \quad Linjun Shou~$^\ddag$ \quad Ming Gong$^\ddag$ \quad Wutao Lin$^\ddag$ \quad Daxin Jiang$^\ddag$\\
 {\small $^\S$ School of Computer Science, Beijing University of Posts and Telecommunications, Beijing, China}\\
  {\small $^\ddag$ STCA NLP Group, Microsoft, Beijing, China} \\
  {\small \tt yangze01@bupt.edu.cn}\\
  {\small \tt \{lisho, migon, wutlin, djiang\}@microsoft.com}
}

\date{}

\begin{document}
\maketitle
\begin{abstract}
Deep pre-training and fine-tuning models (like BERT, OpenAI GPT) have demonstrated excellent results in question answering areas. However, due to the sheer amount of model parameters, the inference speed of these models is very slow. How to apply these complex models to real business scenarios becomes a challenging but practical problem. Previous works often leverage model compression approaches to resolve this problem. However, these methods usually induce information loss during the model compression procedure, leading to incomparable results between compressed model and the original model. To tackle this challenge, we propose a Multi-task Knowledge Distillation Model (MKDM for short) for web-scale Question Answering system, by distilling knowledge from multiple teacher models to a light-weight student model. In this way, more generalized knowledge can be transferred. The experiment results show that our method can significantly outperform the baseline methods and even achieve comparable results with the original teacher models, along with significant speedup of model inference.

\end{abstract}



\input{paper_body}


\bibliography{paper_reference}
\bibliographystyle{acl_natbib}

\end{document}

%% file: paper_body.tex
\section{Introduction}

Question Answering relevance task is a fundamental task in Q\&A system \cite{cimiano2014ontology} (Table~\ref{t:example} shows an example), which is to distinguish whether an answer could well address the given question. This task can provide a more natural way to retrieve information, and help users find answers more efficiently. 

\begin{table}[htbp]
\small
    \renewcommand{\arraystretch}{1.5}
    \caption{An example of Q\&A Relevance Task.}
    \label{t:example}
    \centering
    \begin{tabular}{lp{5cm}p{7cm}}
    \hline
    \textbf{Question}: &\emph{Can CT scan detect polyps?}\\
    \hline
    \textbf{Passage}: &\emph{Polyps are diagnosed by either looking at the colon lining directly  (colonoscopy) or by a specialized CT scan called CT colography (also called a virtual colonoscopy). Barium enema x-rays have been used in the past and may be appropriate ...} \\ \hline
    \textbf{Label}: &\emph{Relevant} \\
    \hline
    \end{tabular}
    \vspace{-14pt}
\end{table}

This task is formalized as a text matching problem~\cite{xue2008retrieval}. Traditional methods usually used vector space models~\cite{salton1975vector, robertson1999okapi}, shallow neural network models~\cite{HuangHGDAH13, ShenHGDM14, PalangiDSGHCSW14} to model the interaction similarity. 

In recent years, deep pre-training approaches \cite{radford2018improving, devlin2018bert} have brought great break-through in NLP tasks. For question answering systems, it also shows very promising results (like QnA relevance, MRC tasks, etc.). However, due to the sheer amount of parameters, model inference is very time-consuming. Even with powerful GPU machines, the speed is still very limited, as shown in Table~\ref{t:example_inference}\footnote{For fair comparison, we set the batch size as 1, and limit the GPU memory as 1GB.}. 

\vspace{-4pt}
\begin{table}[htbp] 
\small
    \caption{The inference speed of BERT on 1080Ti GPU.}
    \label{t:example_inference}
    \centering
    \begin{tabular}{@{}cc@{}}
        \toprule
        \textbf{Model Name} & \textbf{Samples Per Second} \\ \midrule
        \textbf{BERT Base} & 52                         \\
        \textbf{BERT Large} & 16                         \\ \bottomrule
    \end{tabular}
    \vspace{-2pt}
\end{table}

In a commercial question answering system, two approaches are adopted for model inference. i) for head and body queries, large-scale batch-mode processing is used to compute answers in offline. For this part, the number of QnA pairs is at the magnitude of 100 billions, ii) for tail queries, online inference is used and the latency requirement is about 10ms. Both approaches require fast model inference speed. Therefore, we have to perform model compression for inference speedup.

A popular method, called knowledge distillation~\cite{hinton2015distilling} has been widely used for model compression, which implements a teacher-student framework to transfer knowledge from complex networks to simple networks by learning the distribution of the teacher model's soft target (the label distribution provided by teacher's output) rather than the golden label. However, since usually model compression induces information loss, i.e. the performance of student model usually cannot reach parity of its teacher model. Is it possible to have compressed models with comparable or even better performance than that of the teacher model?

To address the above challenge, we may consider an ensemble approach. In other words, we first train multiple teacher models, and then for each teacher model, a separate student model is compressed. Finally the student models ensemble is treated as the final model. Although this approach performs better than the single teacher approach, it takes more capacity due to multiple student models. If we compare the ensemble model with the teacher model, we are actually using capacity to trade off speed. And if we compare this ensemble approach with the single student model, the reason why it is better is as the following. Each teacher may over-fit the training data somehow. If we have multiple teachers, the ensemble approach can cancel off the over-fitting effect to certain degree. However, the over-fitting bias has been transferred from the teacher to the student during the distillation process. The cancelling off is like ``late calibration''. Can we do ``early calibration'' during the distillation stage? 


Based on the above motivations, we propose a unified \textbf{M}ulti-task \textbf{K}nowledge \textbf{D}istillation \textbf{M}odel (\textbf{MKDM} for short) for model compression. Specifically, we train multiple teacher models to obtain knowledge, then we design a multi-task framework to train a single student by leveraging multiple teachers' knowledge, hence improve the generalization performance and cancel off the over-fitting bias during the distillation stage.

The major contributions of our work are summarized as follows:
\begin{itemize}[itemsep= -0.2em,topsep = 0.4em, align=left, labelsep=-0.2mm]
    \item We design a multi-task learning paradigm to jointly learn multiple teacher's knowledge, thus our model can improve the generalization performance by leveraging the knowledge complementary among different teachers.
    \item We make the first attempt to investigate the effective training of Multi-task knowledge distillation for model compression, and also explore a Two Stage Multi-task Knowledge Distillation Model for web-scale Question Answering system.
    \item We conduct experiments on large scale datasets for business scenario to verify the effectiveness of our proposed approach compared with different baseline methods. 
\end{itemize}

The rest of the paper is organized as follows. After a summary of related work in Section 2, we describe the overall design of \textbf{MKDM} in Section 3. Then we describe our proposed model in details in Section 4. We conduct experiments for comprehensive evaluations in Section 5. Finally, section 6 concludes this paper and discuss future directions.
\begin{figure*}[t!]
    \centering
    \includegraphics[scale=0.7, viewport=175 180 625 445, clip=true]{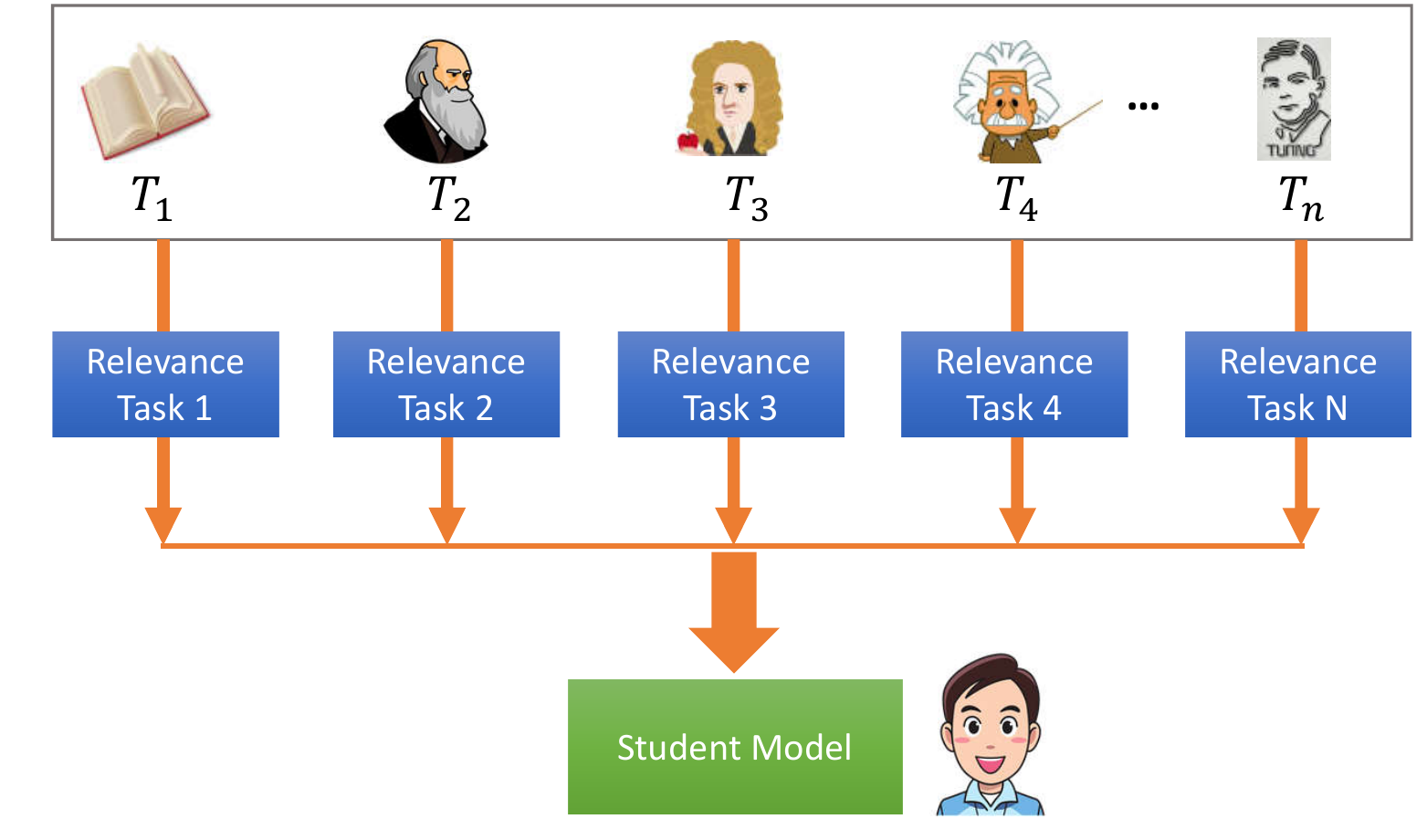}
    \vspace{-10pt}
    \caption{\label{fig:model} The Overall Architecture of The Proposed Multi-task Knowledge Distillation Model.}
    \vspace{-5pt}
\end{figure*}

\section{Related Work}
In this section we briefly review two research areas related to our work: transfer learning and model compression.
\subsection{Transfer Learning}
Transfer learning is a method to transfer knowledge from one task to another, which has been widely used in various fields~\cite{LiZZ16e, MurezKKRK18, BansalSSCD18, he2016deep}. For example, \citeauthor{Cao000L18} \shortcite{Cao000L18} proposed a novel adversarial transfer learning framework to make full use of task-shared boundaries information. \citeauthor{LvCLY18} \shortcite{LvCLY18} proposed a learning-to-rank based mutual promotion procedure to incrementally optimize the classifiers based on the unlabeled data in the target domain. \citeauthor{PengXWPGHT16} \shortcite{PengXWPGHT16} transferred the view-invariant representation of persons' appearance from the source labeled dataset to the unlabeled target dataset by dictionary learning mechanisms.

In recent years, transfer learning has achieved amazing performance on many NLP tasks~\cite{RuderH18,PetersNIGCLZ18,devlin2018bert}. These methods leveraged general-domain pre-training and novel fine-tuning techniques to prevent over-fitting even with small amount of labeled data and achieve state-of-the-art results. In Q\&A system, these methods provide significant improvements. However, These pretrain-finetuning methods need large computation cost due to the large model size.

\subsection{Model Compression}
As the size of neural network parameters is getting larger and larger, how to industrially deploy and apply the model becomes an important problem. Low-rank approximation was a factorization method~\cite{ZhangZMHS15,jaderberg2014speeding,DentonZBLF14}, which used multiple low rank matrices to approximate the original matrix. The main idea of network pruning was to remove the relatively unimportant weights in the network, and then finetune the network~\cite{CunDS89,NIPS1992_647,HeZS17}. \citeauthor{hinton2015distilling} \shortcite{hinton2015distilling} proposed a knowledge distillation method (KD for short) for model compression. In their work, the output of the complex network is used as a soft target for the training of simple network. In this way, the knowledge of complex models can be transferred to simple models. \citeauthor{abs-1802-05668} \shortcite{abs-1802-05668} proposed a quantized distillation method. In their work, they incorporated distillation loss, and expressed with respect to the teacher network, into the training of a smaller student network whose weights are quantized to a limited set of levels. 

Our proposed method is also a knowledge distillation based method. We use a multi-task paradigm to joint learn different teachers' knowledge, and distill the knowledge into a light-weight student.

\section{The Overall Design of Our Model}


Figure~\ref{fig:model} shows the core idea of \textbf{MKDM}. It leverages multiple teachers to jointly train a single student in a unified framework. Firstly, several teacher models are trained using different hyper-parameters. Then we leverage these teacher models to predict soft labels on the training data. Thus each case in training data contains two parts: golden label (the ground truth label in $\{0, 1\}$ for an instance) by human judges and multiple soft labels (the soft labels in $[0, 1]$ predicted by different teacher models). At the training stage, the student model with multiple headers jointly learns the golden label and soft labels. At the inference stage, the final output is a weighted aggregation of all the student headers' outputs. The intuition is very similar to the learning process of human being, i.e. human not only learn knowledge from single teacher, but learn from multiple teachers simultaneously. Unbiased and generalized knowledge can be gained from different teachers.

\section{Our Approach}
In this section, we first describe the proposed approach \textbf{MKDM} in detail\footnote{the code will be released soon.}, and then discuss the model training and prediction details. 

\subsection{\textbf{MKDM}}
\textbf{MKDM} is implemented from BERT~\cite{devlin2018bert}. Our model consists of three layers: the encoder layer utilizes the lexicon to embed both the question and passage into a low embedding space; Transformer layer maps the lexicon embedding to contextual embedding; The multi-task student layer jointly learns multiple teachers' together, and generate prediction output. 

\subsubsection{Encoder Layer}
In Q\&A system, each question and passage are described by a set of words. We take the word piece as the input just like BERT. $X=\{x^{(1)}, x^{(2)}, ..., x^{(|X|)}\}$ is to denote all the instances, and each instance has a $\left \langle Q, P \right \rangle$ pair. Let $Q={\{w_1, w_2, w_3, ..., w_m\}}$ be a question with $m$ word pieces, $P={\{w_1, w_2, w_3, ..., w_n\}}$ be a passage with $n$ word pieces, and $w_i$ is the bag-of-word representation of $i$-th word piece. Each token representation is constructed by the sum of the corresponding token, segment and position embeddings. Let $\textbf{V}^I=\{\vec{v}^I_t\in \mathbb{R}^{D_v}| t=1, \dots, N\}$ denote all the summed vectors in a continuous space.

We concatenate the $\left \langle Q, P \right \rangle$, and $\left \langle CLS \right \rangle$ as the first token, and add $\left \langle SEP \right \rangle$ between Q and P. After that, we obtain the concatenation input $x_c = {\{w_1, w_2, w_3, \dots, w_{m+n+2}\}}$ of a given instance $x^{(i)}$. With the encoder layer, we map $x_c$ into continuous representations $H_e = {\{v_1, v_2, \dots, v_{m+n+2}\}}$. 

\subsubsection{Transformer Layer}
We also use the bidirectional transformer encoder to map the lexicon embedding $H_e$ into a sequence of continuous contextual embedding $H_s = {\{h_1, h_2, h_3, \dots, h_{m+n+2}\}}$. Different from the original BERT, to compress the model, we use a three-layer transformer blocks instead. 

\subsubsection{Multi-task Student Layer}
To jointly learn the multiple teacher models, we design a multi-task layer. In our model, a Teacher Model Zoo is built with different hyper model parameters.

Our multi-task student layer consists of two parts, golden label task and soft label task:
\paragraph{Golden Label Task} Given instance $\left \langle Q, P \right \rangle$, this task aims to learn the ground truth label. Following the BERT, we select $x^{(i)}$'s first token's transformer hidden state $h_1$ as the global representation of input. The probability that $x^{(i)}$ is labeled as class $c$ is defined as follows:
\begin{equation}
    P(c|\left \langle Q, P \right \rangle) = softmax(W^{T}_g\cdot h_1)
\end{equation}
where $W^{T}_g$ is a learnable parameter matrix, $c \in \{0,1\}$ indicates whether $\left \langle Q, P \right \rangle$ is relevant or not. The objective function of golden label task is then defined as the cross-entropy:
\vspace{-5pt}
\begin{equation}\label{eq:gold loss}
    l_g = -\sum\limits_{c \in \{0, 1\}}c \cdot log(P(c|\left \langle Q, P \right \rangle))
\end{equation}

\paragraph{Soft Label Task} For a given instance $\left \langle Q, P \right \rangle$, teacher model can predict a score to indicate the probability that $Q$ and $P$ are relevant. Take a teacher as example, the relevance probability of $\left \langle Q, P \right \rangle$ is defined as follows:
\begin{equation}
    R(Q, P) = sigmoid(W^{T}_s\cdot h_1)
\end{equation} 
where $W^{T}_s$ is a learnable parameter matrix, $R(Q, P) \in [0, 1]$ is the relevance score. 

The objective function of soft label task is defined as mean squared error as follows:
\begin{equation}\label{eq:score_loss}
    l_s = (z - R(P, Q))^2
\end{equation}
where $z$ is the predicted score of teacher for given $\left \langle P, Q \right \rangle$ pairs.

\subsection{Training and Prediction}
\label{sec:train_and_prediction}
In order to learn parameters of \textbf{MKDM} model, we combine Equation~(\ref{eq:gold loss}) and Equation~(\ref{eq:score_loss}), and obtain our multi-task learning objective function as follows:
\begin{equation}\label{eq:multi}
    l = (1-\alpha)l_g + \alpha \frac{1}{n}\sum\limits_{i=1}^{n}l_{s_{i}}
\end{equation}
where $\alpha$ is a loss weighted ratio, $l_{s_{i}}$ is the loss of $i$-th teacher. Details of our learning algorithm is shown in Algorithm (1).
\begin{algorithm}[htbp]
    \caption{Framework of \textbf{MKDM}}
    \label{alg:algo}
    \begin{algorithmic}[1]
        \State Initialize model $\Theta=\{\textbf{W}_g, \textbf{W}_s, \textbf{V}^I$\}
        \label{code:fram:initialize parameters}
        \State iter = 0
        \label{code:fram:begin}
        \Repeat
        \label{code:fram:repeat}
        \State $iter \leftarrow iter+1$
        \label{code:fram:re_value}
            \For{$i = 1,...,|X|$}
            \label{code:fram:for1}
                \State for instance $x^{(k)}$, compute the gradient $\nabla(\theta)$ using Equation (\ref{eq:multi})
                \label{code:fram:com_grad}
                \State update model $\theta \leftarrow \theta + \epsilon \nabla(\theta)$
                \label{code:fram:update_grad}
            \EndFor
        \Until{Converge}
        \State
        \Return $\{\textbf{W}_g, \textbf{W}_s,\textbf{V}^I\}$;
    \end{algorithmic}
    \vspace{-4pt}
\end{algorithm}
\begin{table*}[t!]
    \centering
    \caption{Statistics of experiment datasets.}
    \label{t:statistic}
    \begin{tabular}{ccccc}
    \hline
    \textbf{Datasets}   & \textbf{Number of Samples} & \textbf{\begin{tabular}[c]{@{}c@{}}Average Question Length \\ (words) \end{tabular}} & \textbf{\begin{tabular}[c]{@{}c@{}}Average Answer Length \\ (words) \end{tabular}}  \\ \hline
    \textbf{DeepQA}     & 1,000,000         & 5.86                    & 43.74                                             \\ \hline
    \end{tabular}
    \vspace{-8pt}
\end{table*}

At the inference stage, we use an aggregate operation to calculate the final result as follows:
\begin{equation}
    O(\left \langle P, Q \right \rangle) = \frac{1}{1 + n}(P(1|\left \langle P, Q \right \rangle)
                                            + \sum\limits_{i=1}^{n}R_i(\left \langle P, Q \right \rangle))
\end{equation}
where $R_i$ represent the $i$-th student header's output.

\section{Experiment}

\subsection{Dataset}
Our experimental dataset (called \textbf{DeepQA}) is randomly sampled from one commercial Q\&A system's large dataset. It contains 1 million Q\&A label data covering various domains, such as health, tech, sports, etc. Each case consists of three parts, i.e. question, passage, and binary label (i.e. 0 or 1) by human judges indicating whether the question can be answered by the passage. The statistics are shown in Table~\ref{t:statistic}. 

\subsection{Evaluation metrics}
We use the following metrics for model performance evaluation:
\begin{itemize}[itemsep= -0.2em,topsep = 0.4em, align=left, labelsep=-0.2mm]
    \item \textbf{Accuracy}: This metric equals to number of correct predictions divided by the total number of samples in test set.
    \item \textbf{Area Under Curve}: This metric is one of the most widely used metrics to evaluate binary classification model performance. It equals to the probability that the classifier will rank a randomly chosen positive example higher than a randomly chosen negative example.
    \item \textbf{Queries Per Second}: This metric indicates the numbers of cases to be processed per second. We use this metric to evaluate model inference speed. 
\end{itemize}

\subsection{Baselines}
We compare our model with several strong baseline models to verify the effectiveness of our proposed approach. All the baseline methods are based on BERT pertained model:
\begin{itemize}[itemsep= -0.2em,topsep = 0.4em, align=left, labelsep=-0.2mm]
    \item \textbf{Original BERT}: We use the BERT base fine-tuning model as one strong baseline, which consists of 12-layer transformer blocks, 768 hidden size, and 12 heads. Several BERT base fine-tuning models are trained using different hyper-parameters. 
    \item \textbf{Single Student Model}: 3 layers BERT base model is selected as the student model architecture and model parameters are initialized using the BERT base model weights. Different with \textbf{MKDM}, this student model learns from one single teacher model using knowledge distillation. The teacher model is the best model from the first baseline, i.e. \textbf{Original BERT} model. 
    \item \textbf{Student Model ensemble}: For each BERT base fine-tuning model from the first baseline, knowledge distillation is used to train a BERT 3 layer student model. We train 3 student models using 3 different teacher models, then these student models are ensembled by simply averaging the output scores.
\end{itemize}
\vspace{4pt}
\begin{table*}[t!]
    \small
    \centering
    \caption{Model Comparison Between our Methods and Baseline Methods. ACC, AUC denote accuracy and area under curve respectively (all AUC/ACC metrics in the table are percentage numbers with \% omitted).}
    \label{t:main}
    \begin{tabular}{@{}ccccc@{}}
    \toprule
    \textbf{Model}                  & \textbf{Inference Speed} & \textbf{Parameters} & \multicolumn{2}{c}{\textbf{Performance}} \\
                                    & \textbf{QPS}             & \textbf{(KB)}                & \textbf{ACC}        & \textbf{AUC}       \\ \midrule
    \textbf{Original BERT Model}    & 52                      & 427,721                      & 80.89               & 88.72              \\
    \textbf{Single Student Model}   & 217                      & 178,506                      & 76.29               & 84.12              \\
    \textbf{Student Model Ensemble (3)} & 217 / 3                      & 178,506 * 3                      & 76.77                 & 84.43            \\
    \textbf{MKDM}                   & 217                      & 178,512                      &\textbf{77.18}               & \textbf{85.14}              \\ \bottomrule
    \end{tabular}
    \vspace{-8pt}
\end{table*}

\subsection{Parameter Settings}


For all baselines and \textbf{MKDM}, we implement on top of the PyTorch implementation of BERT\footnote{https://github.com/huggingface/pytorch-pretrained-BERT.}. We optimize \textbf{MKDM} with a learning rate of $3e^{-5}$ and a batch size of 256. In all cases, the hidden size is set as 768. The number of self-attention heads is set as 12, and the feed-forward/filter size is set to 3072.

To compress original BERT model, we set the number of transformer blocks as 3. Teacher models in \textbf{MKDM} are identical to the teacher models of student ensemble model. All baselines and \textbf{MKDM} are not trained from scratch. We finetune the student models on pretrained BERT model weights.

\subsection{Comparison Against Baselines}
In this section, we conduct experiments to compare \textbf{MKDM} with baselines in three dimensions, i.e. inference speed, parameter size and performance. From the results shown in Table~\ref{t:main}, it's intuitive to have the following observations:
\begin{itemize}[itemsep= -0.2em,topsep = 0.4em, align=left, labelsep=-0.2mm]
    \item It's not surprising that original BERT model shows the best performance due to the sheer amount of parameters, but the inference speed is super slow and the memory consumption is huge. 
    \item Single student model obtains pretty good results regarding inference speed and memory capacity, but there are still some gaps compared to the original BERT model in terms of ACC, AUC. 
    \item Student model ensemble performs better than single student model. However, the inference speed and memory consumption increase in proportion to the number of student models ensembled. 
    \item Compared with single student model and student ensemble model, our \textbf{MKDM} achieves optimum in all three dimensions. Compared to the single student model, \textbf{MKDM} only needs small amount of additional memory consumption since majority of the parameters are shared across different tasks. 
\end{itemize}

To conclude, \textbf{MKDM} performs better in three dimensions than two strong baseline compressed models with knowledge distillation (i.e. single student model, student ensemble model) on DeepQA dataset, and also further decreases performance gap with the original BERT model, which verifies the effectiveness of \textbf{MKDM}. 

\subsection{Effective Training of \textbf{MKDM} Model}
In this section, we perform further analysis about how to train \textbf{MKDM} more effectively.

\subsubsection{The Impact of Pre-training Weights}
BERT shows excellent results on plenty of NLP tasks by leveraging large amount of unsupervised data for pre-training to get better contextual representations. In \textbf{MKDM} model, our best practice is leveraging BERT pre-training weights to initialize the first three layers. 

The results in Table~\ref{t:load_pretrain}  show the performance comparison between initializing with pre-training weights and random initializing. 
\begin{table}[htbp]
\small
    \centering
    \caption{\label{t:load_pretrain} The Impact of BERT Pre-training Weights.}
    \begin{tabular}{@{}ccc@{}}
        \toprule
        \textbf{Strategy}                  & \multicolumn{2}{c}{\textbf{Performance}} \\
                                           & \textbf{ACC}      & \textbf{AUC}     \\ \midrule
        \textbf{Random Initializing Weights}              & 67.32             & 77.18            \\
        \textbf{Load Pre-training Weights} & 77.18             & 85.14            \\ \bottomrule
        \end{tabular}
        \vspace{-8pt}
\end{table}

From the results, we can see that model initialized from pre-training weights outperforms training from scratch. The relative performance improvement over ACC, AUC is around $9.86\%$ and $12.55\%$ which is significant. Meanwhile, during the training stage, the pre-training weights makes the model faster to converge.
\begin{figure*}[t!]
    \centering
    \includegraphics[scale=0.7, viewport=125 130 665 470, clip=true]{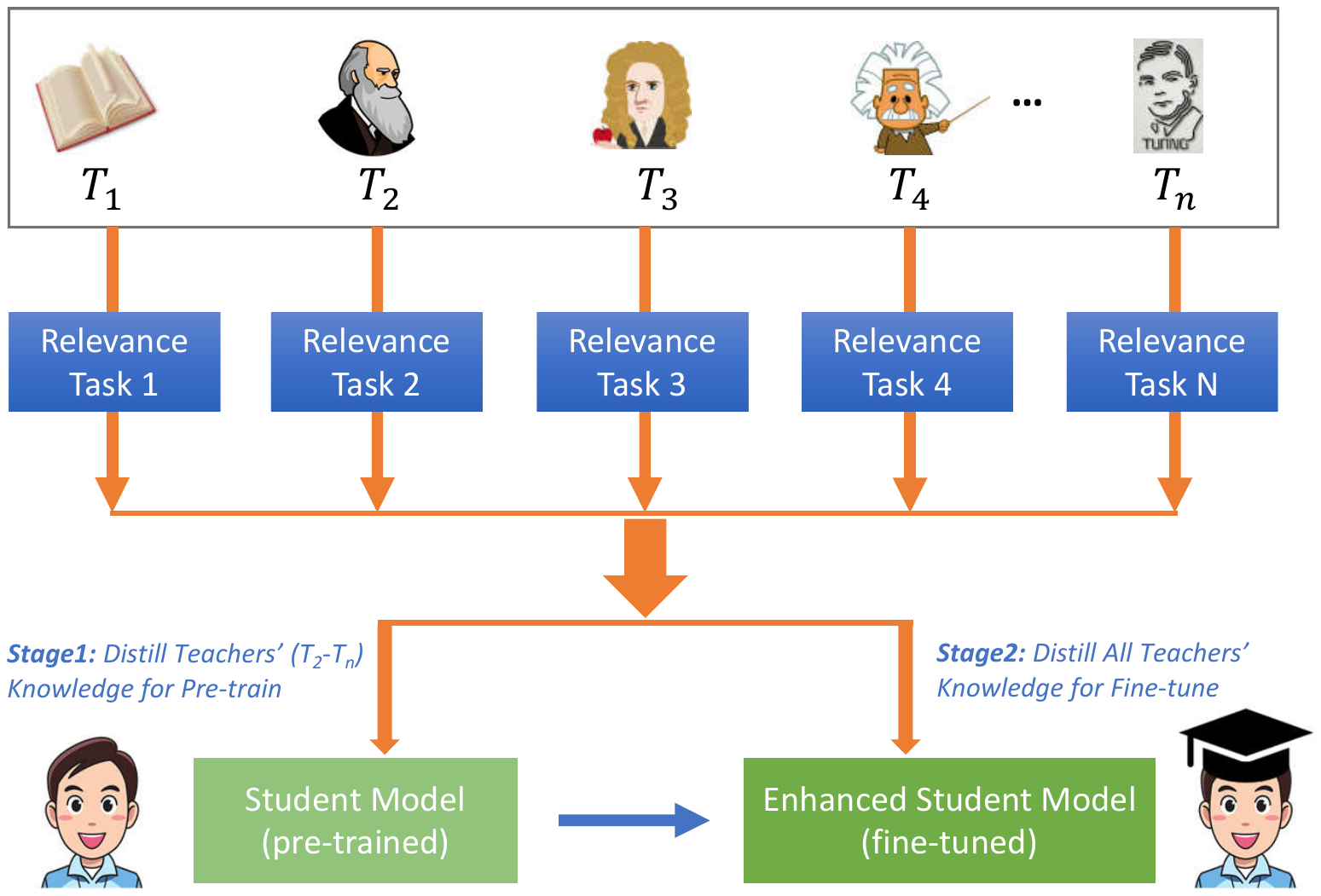}
    \vspace{-10pt}
    \caption{\label{fig:two_stage_model} The Overall Architecture of Our Two Stage Model.}
    \vspace{-5pt}
\end{figure*}

\subsubsection{The Impact of Different Transformer Layer Number}
The most important architecture of BERT is the transformer block count. In \textbf{MKDM}, the number of transformer layer is set as 3. Here we investigate the impact of different numbers of transformer layer. The performance of \textbf{MKDM} is compared when the number of transformer layer $n \in \{1, 3, 5, 7, 9\}$, and the results are shown in Table~\ref{t:different_layer}.
\begin{table}[htbp]
\small
    \centering
    \caption{\label{t:different_layer} The Comparison for Different Number of Transformer Layer.}
    \begin{tabular}{@{}cccc@{}}
    \toprule
    \textbf{Layer Count} & {\textbf{Inference Speed}}   & \multicolumn{2}{c}{\textbf{Performance}}                            \\ 
                          & QPS                                     & \multicolumn{1}{c}{\textbf{ACC}}    &       \multicolumn{1}{c}{\textbf{AUC}} \\ \midrule
    \textbf{1}            &  511                                & 70.02                              & 75.75                              \\
    \textbf{3}            &  217                                & 77.18                              & 85.14                              \\
    \textbf{5}            &  141                                & 78.51                              & 86.65                              \\
    \textbf{7}            &  96                                 & 79.82                              & 87.84                              \\
    \textbf{9}            &  66                                 & 80.57                              & 88.31                              \\ \bottomrule 
    \end{tabular}
    \vspace{-12pt}
\end{table}

From the results, we can draw the following observations:
\begin{itemize}[itemsep= -0.2em,topsep = 0.4em, align=left, labelsep=-0.2mm]
    \item As the number of transformer layer $n$ increases, the AUC and ACC metrics increase as well, but the inference speed decreases. It's easy to understand that more transformer layers bring in larger parameter size which could benefit feature representation for performance, but greatly hurt inference efficiency.
    \item As the number of transformer layer increases, the performance gain between two consecutive trials decreases. That say, when layer count increases from $1$ to $3$, the performance gain over ACC, AUC is $7.16\%$ and $9.39\%$ which is very huge improvement; while increases from $3$ to $5$, $5$ to $7$, $7$ to $9$, the performance gains decrease from around $1.4\%$ to $0.5\%$. Our thinking is that when transformer layer reaches a certain number, the model representation capability seems sufficient and there is no significant add-on value when add more layers.
\end{itemize}

 Based on these results, we set the number of transformer layers as 3 for \textbf{MKDM}, since this setting has the highest performance/computation cost ratio which better meets the requirement for web-scale applications. 
 
 More interestingly, in real business scenario, as the data scale increases, the 3-layer \textbf{MKDM} also shows the potential to achieve comparable results with the original teacher models, which will be introduced in Section~\ref{sec:two_stage_model}.  

\subsubsection{The Impact of Loss Weighted Ratio}
Here we investigate the impact of the loss weighted ratio $\alpha$ defined in Section~\ref{sec:train_and_prediction}, where $\alpha \in \{0.1, 0.3, 0.5, 0.7, 0.9, 1.0\}$. Specially, when set the ratio as $1.0$, we only use the soft label headers to calculate the final output result. Table~\ref{t:different_ratio} shows the performance of \textbf{MKDM} against different $\alpha$ value.
\begin{table}[htbp]
\small
    \centering
    \caption{\label{t:different_ratio} The Impact of Different Loss Weighted Ratio. }
    \begin{tabular}{@{}ccc@{}}
        \toprule
    \textbf{Loss Weighted Ratio} & \multicolumn{2}{c}{\textbf{Performance}} \\
                                & \textbf{ACC}      & \textbf{AUC}     \\ \midrule
        \textbf{0.1}            & 75.48             & 83.39            \\
        \textbf{0.3}            & 75.73             & 83.75            \\
        \textbf{0.5}            & 76.25             & 84.10            \\
        \textbf{0.7}            & 76.56             & 84.38            \\
        \textbf{0.9}            & \textbf{77.18}    & \textbf{85.14}   \\ 
        \textbf{1.0}            & 76.30    & 84.33   \\ \bottomrule
        \end{tabular}
        \vspace{-4pt}
\end{table}

From the results, we obtain the following observations: 
\begin{itemize}[itemsep= -0.2em,topsep = 0.4em, align=left, labelsep=-0.2mm]
  \item The larger ratio, the better performance will be obtained (except when $\alpha$ is $1.0$).
  \item Without the golden label (i.e. $\alpha$ is $1.0$), the performance decreases. Just like when human beings learn knowledge, we let him/her only learn from teachers but without reading any books. Obviously, In this case, they can't master comprehensive knowledge.
\end{itemize}

\subsection{Enhanced Student Model with Two-Stage Multi-Task Knowledge Distillation}
\label{sec:two_stage_model}
In most real business scenarios, it is relatively easy to get large amount of unlabeled $\left \langle Q, P \right \rangle$ data. In \textbf{MKDM}, we only leverage labeled data for model training. In fact, based on \textbf{MKDM} paradigm, we can leverage not only labeled data for knowledge distillation, but also large amount of unlabeled data. Based on this idea, we further propose a \textbf{T}wo \textbf{S}tage \textbf{MKDM} (\textbf{TS-MKDM} for short) approach (as shown in Figure~\ref{fig:two_stage_model}):
\begin{enumerate}[itemsep= -0.2em,topsep = 0.4em, align=left, labelsep=-0.2mm]
    \item Multi-Task Knowledge Distillation for \textbf{pre-training}. That say, at the first stage, student model learns from teacher models' soft labels as the optimization objective.
    \item Multi-Task Knowledge Distillation for \textbf{fine-tuning}. That say, at the second stage, just as the original \textbf{MKDM} model, student model jointly learns the golden label and teacher models' soft labels. 
\end{enumerate}

To verify our idea, we collect two larger commercial datasets (called CommQA-Unlabeled and CommQA-Labeled):
\begin{itemize}[itemsep= -0.2em,topsep = 0.4em, align=left, labelsep=-0.2mm]
    \item CommQA-Unlabeled: It includes around 100 million $\left \langle Q, P \right \rangle$ pairs collected from a commercial search engine (without labels). Firstly, for each question, top 10 relevant documents returned by the search engine are selected to form $\left \langle Question, Url\right \rangle$ pairs; Then passages are further extracted from these documents to form $\left \langle Question, Url, Passage\right \rangle$ triples; Finally $\left \langle Question, Passage \right \rangle$ pairs are used as experimental dataset.
    \item CommQA-Labeled: It is a human labeled dataset, which is several times larger than DeepQA with more diversified data.
\end{itemize}


CommQA-unlabeled is used for the above pre-training stage, DeepQA and CommQA-Labeled are used in above fine-tuning stage respectively to evaluate the performance of \textbf{TS-MKDM}. Table~\ref{t:two_stage_model} shows the comparison results between \textbf{MKDM} and \textbf{TS-MKDM}. From the results, we can observe the following findings:
\begin{itemize}[itemsep= -0.2em,topsep = 0.4em, align=left, labelsep=-0.2mm]
    \item On both datasets, \textbf{TS-MKDM} outperforms \textbf{MKDM} by large margin, which proves that incorporating multi-task knowledge distillation for pre-training can further boost model performance. 
    \item Interestingly, by leveraging multi-task knowledge distillation on super large scale dataset for pre-training, the evaluation results on CommQA-Labeled dataset show that \textbf{TS-MKDM} model even exceeds the performance of teacher model (AUC 87.5 vs 86.5, ACC 79.22 vs 77.00). This further verifies \textbf{TS-MKDM}'s effectiveness.
\end{itemize}

\begin{table}[htbp]
\small
    \centering
    \caption{\label{t:two_stage_model} The Performance comparison between \textbf{MKDM} and \textbf{TS-MKDM}. }
    \renewcommand{\arraystretch}{1.15}
\begin{tabular}{@{}ccccc@{}}
\toprule
\textbf{Dataset}       & \multicolumn{2}{c}{\textbf{DeepQA}} & \multicolumn{2}{c}{\textbf{CommQA-Labeled}} \\
\textbf{Model}         & \textbf{ACC}     & \textbf{AUC}     & \textbf{ACC}                             & \textbf{AUC}                             \\ \midrule
\textbf{Original BERT} & 80.89            & 88.72            & 77.00                                    & 86.50                                    \\
\textbf{MKDM}          & 77.18            & 85.14            & 77.32                                    & 85.71                                    \\
\textbf{TS-MKDM}           & \textbf{78.47}            & \textbf{86.36}            & \textbf{79.22}                                    & \textbf{87.50}                                    \\ \bottomrule
\end{tabular}    
\end{table}

\section{Conclusion and Future Work}
In this paper, we propose a novel Multi-task Knowledge Distillation Model (\textbf{MKDM}) for model compression. A new multi-task paradigm is designed to jointly learn from multiple teacher models. Based on this method, our student model can learn more generalized knowledge from different teachers. Results show that our proposed method outperforms the baseline methods by great margin, along with significant speedup of model inference. We further perform extensive experiments to explore a Two Stage Multi-task Knowledge Distillation Model (\textbf{TS-MKDM}) based on \textbf{MKDM}. The result shows that in real industry scenario with super large scale data, \textbf{TS-MKDM} even outperforms the original teacher model.

In the future, on one side, we will investigate on heterogeneous student models (not transformer based models) to evaluate our multi-task knowledge distillation approach and further boost model agility. On the other side, we will extend our methods to more tasks, such like sentence classification, machine reading comprehension, etc.